\documentclass{article}

\usepackage[preprint, nonatbib]{Styles/neurips_2024}

\usepackage[utf8]{inputenc} 
\usepackage[T1]{fontenc}    
\usepackage{hyperref}       
\usepackage{url}            
\usepackage{booktabs}       
\usepackage{amsfonts}       
\usepackage{nicefrac}       
\usepackage{microtype}      
\usepackage{xcolor}         

\usepackage[normalem]{ulem}
\usepackage{amsmath,amssymb,amsfonts}
\usepackage{algorithm}
\usepackage{algpseudocode}
\usepackage{graphicx}
\usepackage{textcomp}
\usepackage[normalem]{ulem}
\usepackage[backend=bibtex, style=numeric]{biblatex} 
\usepackage{wrapfig}

\title{ENDOR: Hardware-Fri\underline{end}ly Sparse Format for \underline{O}ffloaded LLM Infe\underline{r}ence}

\author{%
    \small \textbf{Donghyeon Joo$^{1}$, Ramyad Hadidi$^{2}$, Soheil Feizi$^{1}$, Bahar Asgari$^{1}$} \\
    \small$^{1}$Department of Computer Science, University of Maryland, $^{2}$Rain AI \\
    \small \texttt{\{dhjoo98,bahar\}@umd.edu}, \texttt{ramyad@rain.ai},  \texttt{sfeizi@cs.umd.edu}
}

\begin{document}

\maketitle

\begin{abstract}
  The increasing size of large language models (LLMs) challenges their usage on resource-constrained platforms. For example, memory on modern GPUs is insufficient to hold LLMs that are hundreds of Gigabytes in size. Offloading is a popular method to escape this constraint by storing weights of an LLM model to host CPU memory and SSD, then loading each weight to GPU before every use. In our case study of offloaded inference, we found that due to the low bandwidth between storage devices and GPU, the latency of transferring large model weights from its offloaded location to GPU memory becomes the critical bottleneck with actual compute taking nearly 0\% of runtime.    
  To effectively reduce the weight transfer latency, we propose a novel sparse format that compresses the unstructured sparse pattern of pruned LLM weights to non-zero values with high compression ratio and low decompression overhead. Endor achieves this by expressing the positions of non-zero elements with a bitmap. Compared to offloaded inference using the popular Huggingface Accelerate, applying Endor accelerates OPT-66B by 1.70$\times$ and Llama2-70B by 1.78$\times$. When direct weight transfer from SSD to GPU is leveraged, Endor achieves 2.25$\times$ speedup on OPT-66B and 2.37$\times$ speedup on Llama2-70B. 

\end{abstract}

\section{Introduction}

Large language models (LLMs) have seen an exponential increase in size to achieve meta-human abilities such as complex reasoning and zero-shot tasks. However, the immense size of LLMs, such as 70-billion-parameter Llama~\cite{llama2} and 175-billion-parameter GPT-3~\cite{gpt3}, pose challenges to the computing platform. LLM inference is often limited by the GPU memory when it cannot hold the entire model. LLM practitioners often face a peculiar situation where the `GPU-must-hold-entire-model' rule limits inference even with enough computing resources. This forces them to employ more GPUs, which is costly and illogical as already-sufficient compute resources are added. 
Model offloading presents a promising alternative. By partitioning a model into small sections and saving them to host CPU memory and storage device, computing platforms can overcome the limitations of GPU memory. Now comfortably fitting on GPU memory, these partitions are sequentially loaded during inference time. This approach, supported by prominent LLM frameworks such as Hugging Face Accelerate~\cite{hfaccelerate} and Microsoft Deepspeed~\cite{deepspeed}, holds great potential for LLM inference. 

In this paper, upon analyzing offloaded inference (Section~\ref{sec:3}), we identify the critical bottleneck as the weight transfer latency of offloaded model weights. This overhead is significantly larger than the time spent in computation, resulting in a much longer end-to-end latency than the on-GPU inference. 
Model quantization has been a popular method in LLM deployment, decreasing the memory footprint by lowering the precision of each value. Alternatively, model pruning can achieve a similar level of model size reduction by removing relatively unimportant weight values with minor performance degradation.
However, pruning methods such as SparseGPT~\cite{sparsegpt}, which yield unstructured sparse pattern in weights do not reduce the size of stored weights due to the difficulty in expressing the unstructured distribution of non-zeros. When pruned, weights are stored as-is, it does not contribute to reducing weight transfer latency. 

To address the bottleneck of offloaded LLM inference with minimal model performance degradation, we propose Endor (Section~\ref{sec:4}), a novel sparse format to efficiently compress the pruned unstructured sparse LLM weights. Endor efficiently compresses unstructured sparse weight matrices to reduce the weight transfer latency in transfer between storage device and host CPU memory, and between host CPU memory and GPU. To achieve high compression rate, Endor sparse format stores only the non-zero elements of the sparse LLM weights and use a bitmap to express the position of these elements. Decompression of Endor sparse format is highly parallelizable and shows minimal decompression overhead. Endor achieves up to 2.37$\times$ speedup compared to the dense offloaded inference using Huggingface Accelerate~\cite{hfaccelerate} (Section~\ref{sec:evaluation}). 
As Endor preserves all non-zero values of pruned weights, it can be jointly applied with other methods that accelerate offloaded inference, such as quantization and activation sparsity for faster offloaded inference, and batch scheduling for higher throughput. 

\section{Related Works}
This section first reviews different weight size reduction methods we considered and then reviews previous studies that target offloaded inference.  
\subsection{\textbf{Model Pruning, Quantization, and Compression Format}} 

Several studies have proposed model pruning~\cite{prune1, prune2, prune3, prune4, prune5} that replaces insignificant elements in a weight tensor with zeros. Applying pruning to LLMs, SparseGPT~\cite{sparsegpt} employs an iterative pruning process with weight updates to decide the values to prune. Wanda~\cite{wanda} uses a magnitude-based sorting of products between weight values and sample dataset inputs. In addition for both methods to prune in unstructured pattern, they can also enforce N:M structured pattern, where N elements remain in M consecutive elements, for computational speedup on supported GPUs~\cite{nmsparsity}. However, enforcing a structured pattern comes at the price of model accuracy degradation.

The same limitation of accuracy degradation apply when considering a structured sparse pattern to reduce the memory overhead of pruned weights. Alternatively, compressing the unstructured sparse pattern is a challenging task due to the random distribution of non-zero elements. If pruned weights are compressed using common sparsity formats such as compressed sparse row (CSR) fromat, the addition indexing data makes up for the reduction of elements. For example, the size of indexing data for a 50\% sparse weight is equal to the size of zero elements, a compression rate of 0\%. This forces pruned weights to be stored as whole dense matrices, with no weight size reduction despite the reduction in the number of weight elements. In our paper, we use weights pruned to 50\% sparsity with the methods of Wanda~\cite{wanda} and SparseGPT~\cite{sparsegpt}, resulting in an unstructured distribution of non-zero values. Endor is applied on the pruned weights to reduce weight size during offloaded inference. 

Model quantization~\cite{quant1, quant2, quant3, quant4, quant5} is an alternative solution that reduces both the model weight size and the amount of computation by reducing the precision of each value. For LLMs, SmoothQuant~\cite{smoothquant} performs INT8 quantization of weights and activations to achieve memory reduction and speedup. GPTQ~\cite{gptq} uses a layer-wise quantization to achieve more reduction with minimal accuracy degradation. SpQR~\cite{sqpr} retains higher precision outlier values in a mixed-precision scheme to minimize accuracy degradation. We design Endor sparse format to be compatible with LLM quantization, as evaluated in Section~\ref{sec:quantization}

State of the art compression formats such as LZ4~\cite{lz4} and Meta Zstandard (ZSTD)~\cite{zstd} encapsulate decades of effort in computing compression formats, which exploit the repetition of bit-wise patterns in raw data to achieve size reduction\footnote{We distinguish compression with compression format, which exploits bit-wise patterns, and compression with sparse format, such as CSR and Endor, that exploits the non-zeros of sparse matrices. Unless stated explicitly, `compression' refers to `compression with sparse format.'}. In Section~\ref{sec:perlayerspeedup}, we evaluate the usage of ZSTD to compress model weights in offloaded inference.

\subsection{\textbf{Offloaded Inference Optimization}}

Alizadeh et al. \cite{llminaflash} exploits the activation sparsity of the ReLU function to load weights from storage selectively. They use a small predictor that accurately predicts the weight rows and columns that yield non-zero ReLU activation and load that subset of weights, leading to a 95\% reduction of weight transfer. However, their target is limited to the fully connected layers of an LLM, which is, on average, 66\% of the entire model weight transfer. Also, only a subset of prominent LLMs such as OPT and Falcon that employs ReLU activation benefit from this approach, excluding LLMs such as Llama and GPT4. 
Sheng et al. \cite{flexgen} employs efficient computation scheduling, tensor placement, and KV cache compression to increase the throughput of offloaded inference. They achieve this by establishing a search space of possible configurations given a batch input of sequences. However, their work prioritizes increasing the throughput of batch inputs, which is suitable for a server-scale LLM serving. We aim to reduce the offloaded inference latency of a single batch input sequence, which is suitable for both server-scale and edge-scale LLM serving. 

\section{Case Study: Offloading OPT-66B}
\label{sec:3}
\subsection{\textbf{Workload Specification}}
\label{sec:case_study}

Our workload is OPT-66B~\cite{opt} in float16 precision. OPT-66B consists of 64 OPT layers. Each OPT layer is divided into an attention sub-layer and a fully-connect sub-layer. Attention sub-layer contains four linear operations: key projection, query projection, value projection, and output projection with weight matrix of size $9,216 \times 9,216$. Fully-connected sub-layer contains two linear operations with weight matrix of size $9,216 \times 36,864$. Linear operations are the focus of our work, as their weight parameters are significantly larger than operations with smaller parameters (layer normalization) and operations with no weight parameter (matrix multiplication between attention score and value). With emphasis on offloaded inference latency, we use single batch text generation inference for all measurement and evaluation. 

\subsection{System Specification}
\begin{figure}[h]
\centerline{\includegraphics[width=0.35\textwidth]{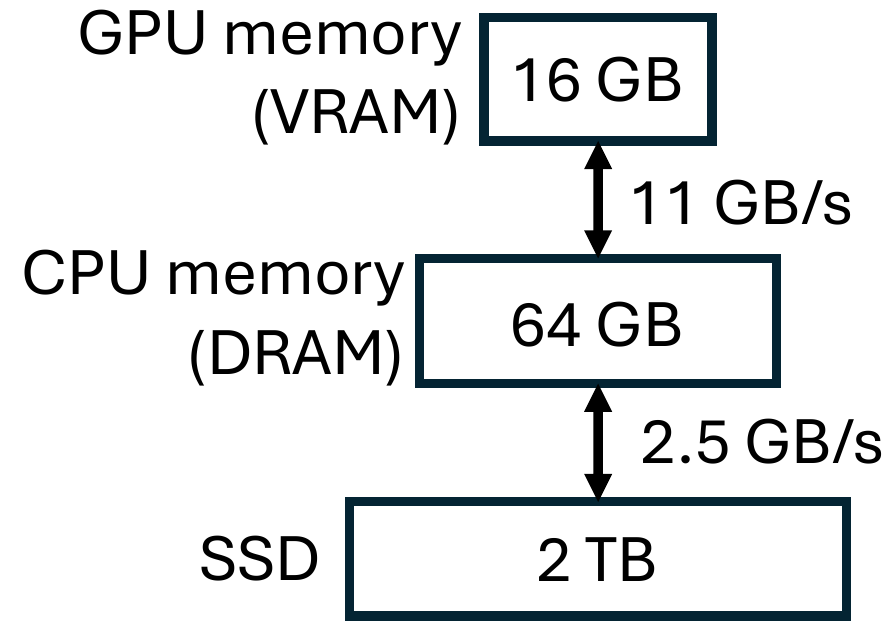}}
\caption{Memory/SSD configuration including capacity and measured bandwidth.\vspace{-20pt}}
\label{fig:capacity_bandwidth}
\end{figure}

Fitting our workload on GPU would require 132GB of VRAM, far surpassing the memory size of commercial GPUs.  We use Hugging Face Accelerate to perform offloaded inference on a CPU-GPU heterogeneous platform, which consists of an RTX 4080 GPU with 16GB VRAM, 64GB host DRAM, and SK Hynix P31 NVMe 2TB SSD.  Figure~\ref{fig:capacity_bandwidth} illustrates the measured bandwidth between these devices. Bandwidth between storage device and host CPU DRAM is significantly smaller compared to the bandwidth between host CPU DRAM and the GPU.

\subsection{Offloading Setup}
Offloading involves mapping each OPT layer to GPU memory, host CPU memory, or storage device. OPT layer, which is the unit of offloading, refers to a group of operations comprised of attention, layer normalization, and fully-connected operations. GPU-mapped-layer weights are directly computed. CPU-mapped-layer weights are transferred to GPU memory before computation, while storage-mapped-layer weights are loaded to CPU and then to GPU before computation. To study offloaded inference, we map decoder layers 0 to 4 to GPU memory, layers 5 to 12 to CPU memory, and layers 13 to 63 to the SSD. Layer mapping prioritizes populating locations closer to GPU while leaving enough memory for computation and intermediate operands.  

\begin{figure}[htbp]
\centerline{\includegraphics[width=0.6\textwidth]{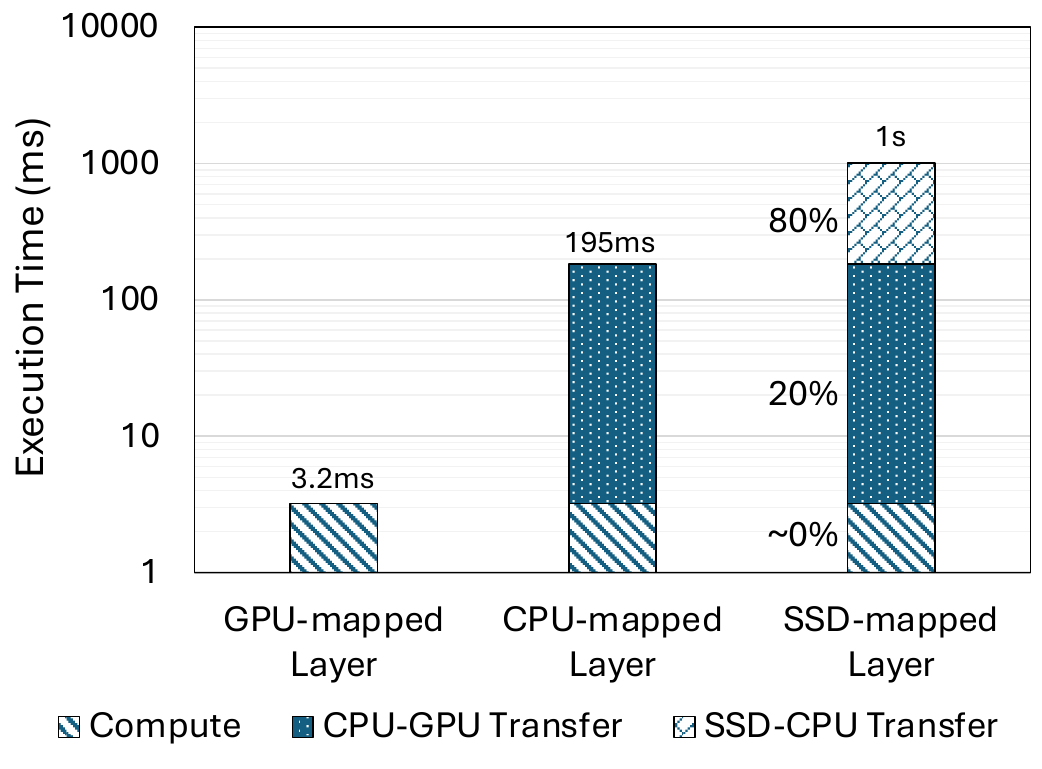}}
\caption{Execution time comparison of offloaded OPT layers.}
\label{fig:execution_time_breakdown}
\end{figure}

\subsection{Offloaded Inference Analysis}
Out of a single pass through OPT-66B which took 54s, Figure~\ref{fig:execution_time_breakdown} compares the execution time of a single GPU-mapped, CPU-mapped, and SSD-mapped layer. The figure illustrates that the overhead of weight transfer increases significantly as the offloaded region is further from the GPU. The SSD-mapped layers have the slowest latency of 1 second, 80\% of which is spent loading weight from storage to CPU, and 20\% is spent in DRAM to GPU loading, resulting in nearly 0\% of time spent in actual computation. For the the CPU-mapped layer, weight only has to be transferred from CPU DRAM to GPU VRAM before computation, and thus does not include SSD to CPU transfer overhead. For the GPU-mapped layer, weight can be used immediately, thus does not include any weight transfer overhead. 
 
\begin{figure}[h]
\centerline{\includegraphics[width=0.70\textwidth]{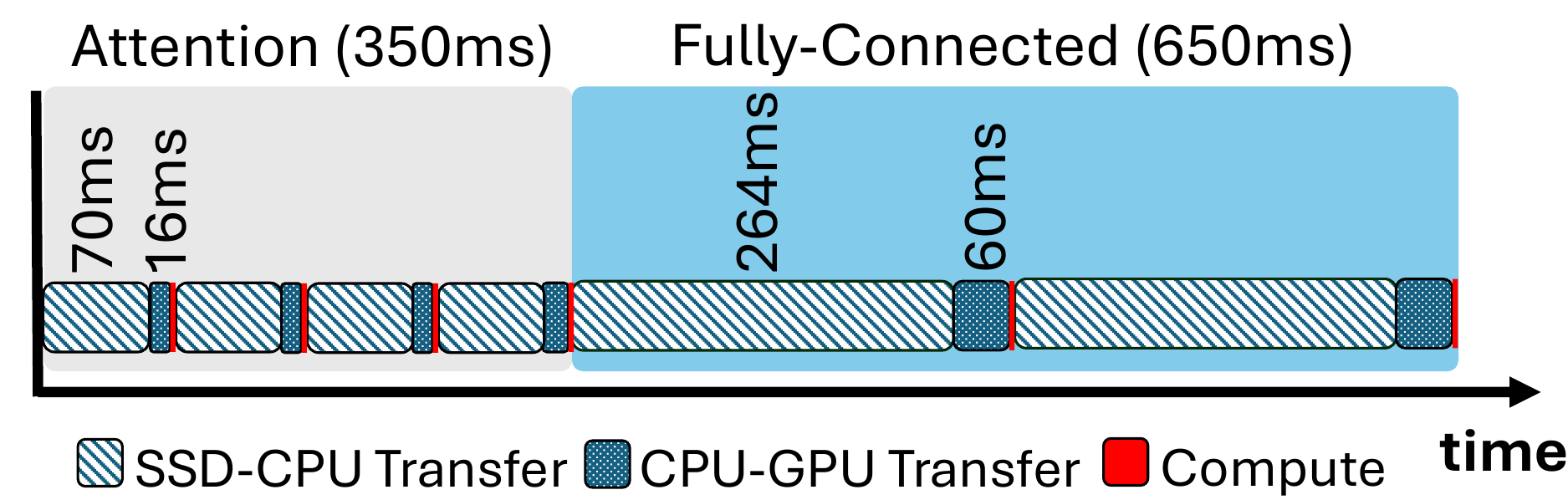}}
\caption{Timeline of an SSD-mapped OPT layer.}
\label{fig:SSD_timeline}
\end{figure}

The impact of SSD-CPU and CPU-GPU transfer times is in particular crucial in the LLM generation task, which involves sequential computations of all layers for every token that requires transferring weights from SSD. To explore this further, Figure~\ref{fig:SSD_timeline} illustrates the timeline for an SSD-mapped OPT layer, showing how weight transfer dominates the execution time. Such a  proportionate weight transfer latency is a result of the small CPU-SSD bandwidth shown in Figure~\ref{fig:capacity_bandwidth}. Some offloading frameworks, such as Deepspeed~\cite{deepspeed}, overlap computation with loading. However, this has little effect as computation takes only a minute portion of execution time, rendering overlapping ineffective. 

\subsection{Overcoming Bottleneck: Unstructured Sparsity Compression}

\label{sec:QorP}
Our two candidates for reducing the identified bottleneck weight transfer latency were quantization and pruning, which can reduce the size of weight parameters. In theory, reducing weight size would lead to a proportionate reduction of weight transfer latency. However, weight pruned in an unstructured pattern must be stored as a whole matrix, and the reduction of weight elements is not reflected in the weight transfer latency. For efficient LLM deployment, this constrains pruning pattern to make a structure, such as row-wise or 2:4 sparsity, which seriously deteriorates model accuracy compared to an unstructured pattern.
Therefore, we propose a novel method to reflect the reduction of weight element to actual reduction in weight size by efficiently compressing the unstructured sparsity pattern. This retains the model accuracy while increasing the practicality and applicability of pruning techniques. We detail our sparse format in Section~\ref{sec:endor_format}.
As the realms of quantization and pruning evolve, achieving good performance with even lower precision or with higher pruning ratio, our work aims to even the comparison ground by providing a way for LLM pruning to yield reduced memory footprint. 
We also design our method with the potential for joint application of pruning and quantization in mind. We explore the Endor's effectiveness in jointly applying quantization and pruning in Section~\ref{sec:quantization}.


\begin{minipage}{0.48\textwidth}
\begin{algorithm}[H]
\caption{The compression algorithm. From a 50\% pruned weight matrix $\mathbf{W}$, we extract the non-zero elements into an array $\mathbf{v}$, and express the position of zeros and non-zeros as bit 0 and 1 into a binary array $\mathbf{b}$.}\label{alg:cap}
\begin{algorithmic}[1]

\State Let $\mathbf{W}$ be a matrix with $m$ rows and $n$ columns
\For{$i = 1$ to $m$}  
    \State Initialize $\mathbf{v}_i$ as an empty list  
    \State Initialize $\mathbf{b}_i$ as an empty list  
    \For{$j = 1$ to $n$}  
        \If{$\mathbf{W}_{i,j} \neq 0$}  
            \State Append $\mathbf{W}_{i,j}$ to $\mathbf{v}_i$  
            \State Append 1 to $\mathbf{b}_i$  
        \Else
            \State Append 0 to $\mathbf{b}_i$  
        \EndIf
    \EndFor
\EndFor
\end{algorithmic}
\end{algorithm}
\end{minipage}
\begin{minipage}{0.48\textwidth}
\begin{algorithm}[H]
\vspace{9pt}
\caption{The decompression algorithm. To occupy a row of decompressed weight matrix $\mathbf{W}$, non-zeros from array $\mathbf{v}$ are sequentially placed in the position of ones in the binary position array $\mathbf{b}$.}\label{alg:decompression}

\begin{algorithmic}[1]
\State Let $\mathbf{W}$ be a matrix with $m$ rows and $n$ columns
\State Assume $\mathbf{v}$ is an array of non-zero elements
\State Assume $\mathbf{b}$ is a binary position array 
\For{$i = 1$ to $m$}  
    \State Initialize $k \gets 0$  
    \For{$j = 1$ to $n$}  
        \If{$\mathbf{b}[j] = 1$}  
            \State $\mathbf{W}_{i,j} \gets \mathbf{v}[k]$  
            \State $k \gets k + 1$  
        \EndIf
    \EndFor
\EndFor
\vspace{4pt}
\end{algorithmic}
\end{algorithm}
\end{minipage}

\section{Endor Sparse Format and Pruning}
\label{sec:4}
\subsection{Endor Sparse Format} 
\label{sec:endor_format}

\begin{figure}[b]
\centerline{\includegraphics[width=0.75\textwidth]{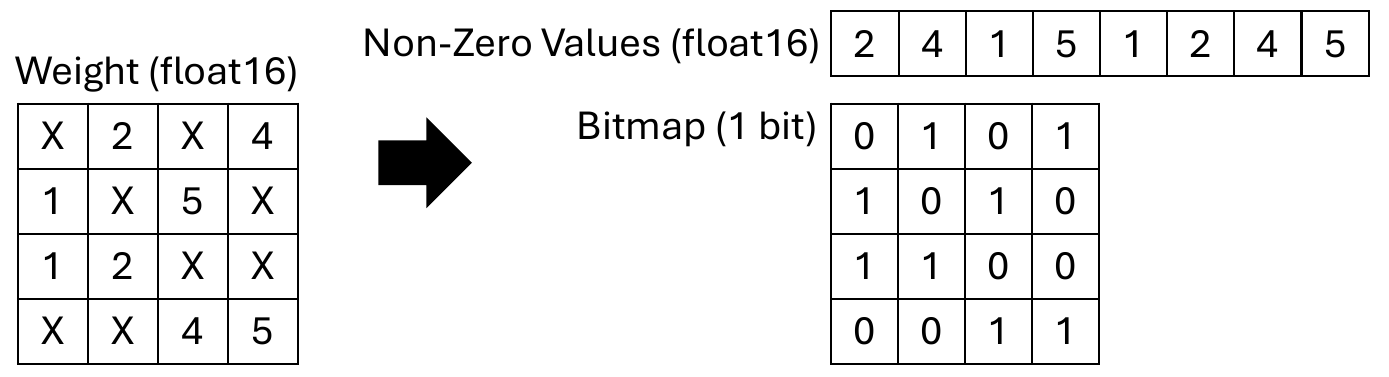}}
\caption{Sparse format for offloaded model weights. Bitmap is stored as a 1-d vector.} 
\label{fig:sparse_format}
\end{figure}

Our proposed method reduces the size of each layer weight, previously unexplored due to the unstructured nature of sparse patterns. As Figure~\ref{fig:sparse_format} shows, for every sparse weight matrix derived from pruning, we save only the non-zero elements and express the location of non-zero elements as a binary array, or a bitmap.  
In terms of compression ratio, the bitmap is a minimal addition to the weight size that is easily amortized by the much larger size reduction of removing zero-valued elements, reaching a compression ratio close to pruning ratio.   
We pruned our workload OPT-66B to 50\% sparsity with pruning algorithm from Wanda~\cite{wanda}. Each fully connected layer weight is a $9,216 \times 36,864$ float16 matrix of size 680MB. Using our sparse format, the isolated non-zero values become 340MB, and the bitmap is a $9,216 \times 36,864$ binary  matrix of size 42MB, reducing the overall weight size to 56\% of the original size. The reduction of weight size leads to a proportionate reduction of weight transfer latency.  

\noindent\textbf{Compression.} Algorithm~\ref{alg:cap} presents the compression of the pruned weight matrix into the proposed sparse format. Compression of weights and their placement into storage devices are performed before inference time; thus, it is irrelevant to inference overhead. 

\noindent\textbf{Decompression.} 
Algorithm~\ref{alg:decompression} presents the decompression of the sparse format. Decompression is performed on the GPU during inference time after the weight in sparse format is loaded from storage to host CPU memory and then from host CPU memory to GPU memory. We found GPU is more suitable to leverage the parallel nature of the decompression algorithm at a lower latency than decompression on CPU. We have studied its effect in detail in Section~\ref{sec:evaluation}.

\subsection{Joint Application with Quantization and Activation Sparsity}
\label{sec:joint_application}

Endor sparse format for pruned LLM is orthogonal to other optimization methods of offloaded inference, mainly quantization and activation sparsity. 
Acknowledging the potential of jointly applying pruning and quantization, Endor reduces the weight transfer latency of pruned LLM regardless of its bit-width. However, the compression rate will decrease. For example, when an 8-bit quantization is applied, a weight matrix of $9,216 \times 36,864$ is naively of size 340MB. Applying Endor sparse format with 50\% pruning would yield the isolated non-zero values to 170MB and the bitmap 42MB. Overall, the compression rate is now 62\% as compared to 56\% in a 16-bit weight matrix. We measure the speedup of quantization with Endor sparse format in Section~\ref{sec:quantization}.
Leveraging activation sparsity involves a small network that predicts which neurons will be non-zero after the ReLU activation function. Using this prediction, the rows of up projection weight matrix and columns of down projection weight matrix is selectively loaded. Because Endor preserves the non-zero values of the pruned weight and expresses its position with a bitmap, a minimal bitmap processing can determine the non-zero values to selectively load.

\section{Evaluation}
\label{sec:evaluation}

We used the LLM model and system configuration from Section~\ref{sec:case_study} for evaluation. Additionally, we measure the effect of Endor on Llama2-70B on Appendix~\ref{sec:appendix_Llama}.

\subsection{\textbf{Compressed Offloaded Inference}}
\label{sec:perlayerspeedup}
To evaluate Endor, we first compare Endor's performance with baseline, decompression on CPU, and ZSTD compression on a single linear operation of OPT. We then compare Endor's effect with baseline on the inference pass of entire OPT-66B.

\begin{figure}[htbp]
\centerline{\includegraphics[width=0.90\textwidth]
{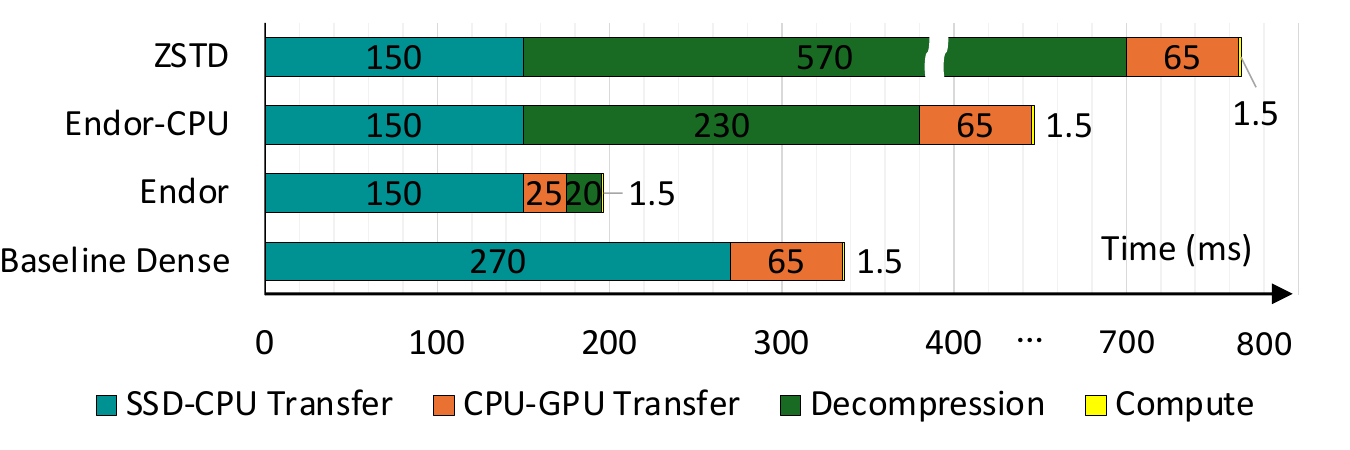}}
\caption{Timeline of offloaded execution of fully-connected operation}
\label{fig:fig5}
\end{figure}

\noindent\textbf{Per-Operation Speedup.}  
Figure~\ref{fig:fig5} shows the offloaded execution of OPT layer's fully-connected operation using our novel sparse format compression. Compared to the baseline offloaded inference using dense weight matrices, this compression method, which significantly reduces the weight size, leads to a 1.67$\times$ speedup of both storage to CPU and CPU to GPU weight transfer. Negligible latency added by decompression on GPU is far outweighed by the reduction of weight transfer overhead. Breakdown of other operations of OPT-66B for each methods are listed in Appendix~\ref{sec:otherops}.

$\bullet$~~\textit{Decompression on CPU vs GPU --} 
We compare the effect of decompression performed on CPU and GPU. On CPU, decompression is parallelized with multiprocessing on 12 CPU cores. Identical with Endor, reduced weight size reduces the dominant weight transfer overhead from storage to CPU. Even with multiprocessing, the decompression overhead overshadowed the weight transfer reduction. Also, because decompression is performed on CPU, there is no reduction in weight transfer overhead from CPU to GPU.

$\bullet$~~\textit{Comparison with ZSTD --}
To compare Endor latency with a compression format, we implemented an offloaded inference that uses offloaded weights compressed with ZSTD. During inference, zstandard executable is used to perform decompression on CPU. As shown on Figure 5, our sparse format compression outperforms ZSTD. While ZSTD achieves a similar compression ratio by exploring bit-level patterns of a pruned weight matrix (i.e., 58\%), the decompression overhead itself exceeds the benefit gained from reduced weight transfer overhead. On the other hand, Endor sparse format can be decompressed much faster. Endor decompression overhead is smaller than just the reduction of CPU-GPU transfer.

\begin{figure}[t]
\centerline{\includegraphics[width=0.95\textwidth]
{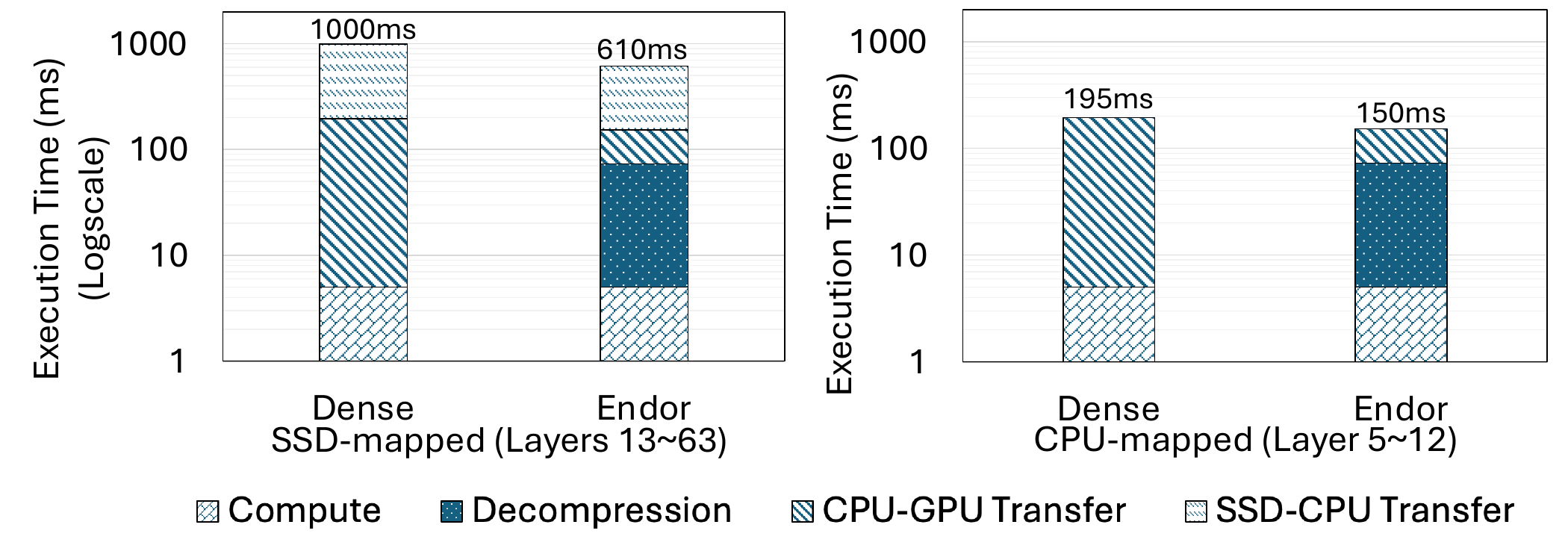}}
\caption{Execution time comparison of dense and Endor offloaded OPT layers for SSD-mapped layers (on left) and CPU-mapped layers (on right).}
\label{fig:fig6}
\end{figure}

\noindent\textbf{End-to-end Speedup.} 
We measure the execution time for single pass through OPT-66B during text generation. For accurate comparison with dense offloaded inference, we keep the same device mapping: Layers 0 to 4 on GPU, layers 5 to 12 on CPU, and layers 13 to 63 on storage. 
For a single pass through OPT-66B, dense offloaded inference takes 54s while Endor offloaded inference takes 35.8s, an overall 1.51$\times$ speedup. Figure~\ref{fig:fig6} breaks down the speedup into each device-mapped layers. In CPU-mapped layers, the reduced overhead of CPU to GPU weight transfer is more than enough to compensate for the added Endor decompression overhead for a 1.30$\times$ speedup. 

SSD-mapped layers see even more speedup of 1.64$\times$ as storage to CPU weight transfer overhead is also reduced by Endor. GPU-mapped layers have the same execution time as in both offloaded inference, as weights are kept in GPU as dense matrices.  
Since each weight size is reduced to 56\% in size, sparse format allows more layers to reside in CPU memory. Compare to 8 OPT layers mapped to CPU in naïve offloaded inference, Endor allows 14 OPT layers to reside in CPU with the same memory consumption. With more layers mapped to CPU, we observe a 1.70$\times$ speedup to baseline offloaded inference.  

\begin{figure}[t]
\centerline{\includegraphics[width=1.0\textwidth]
{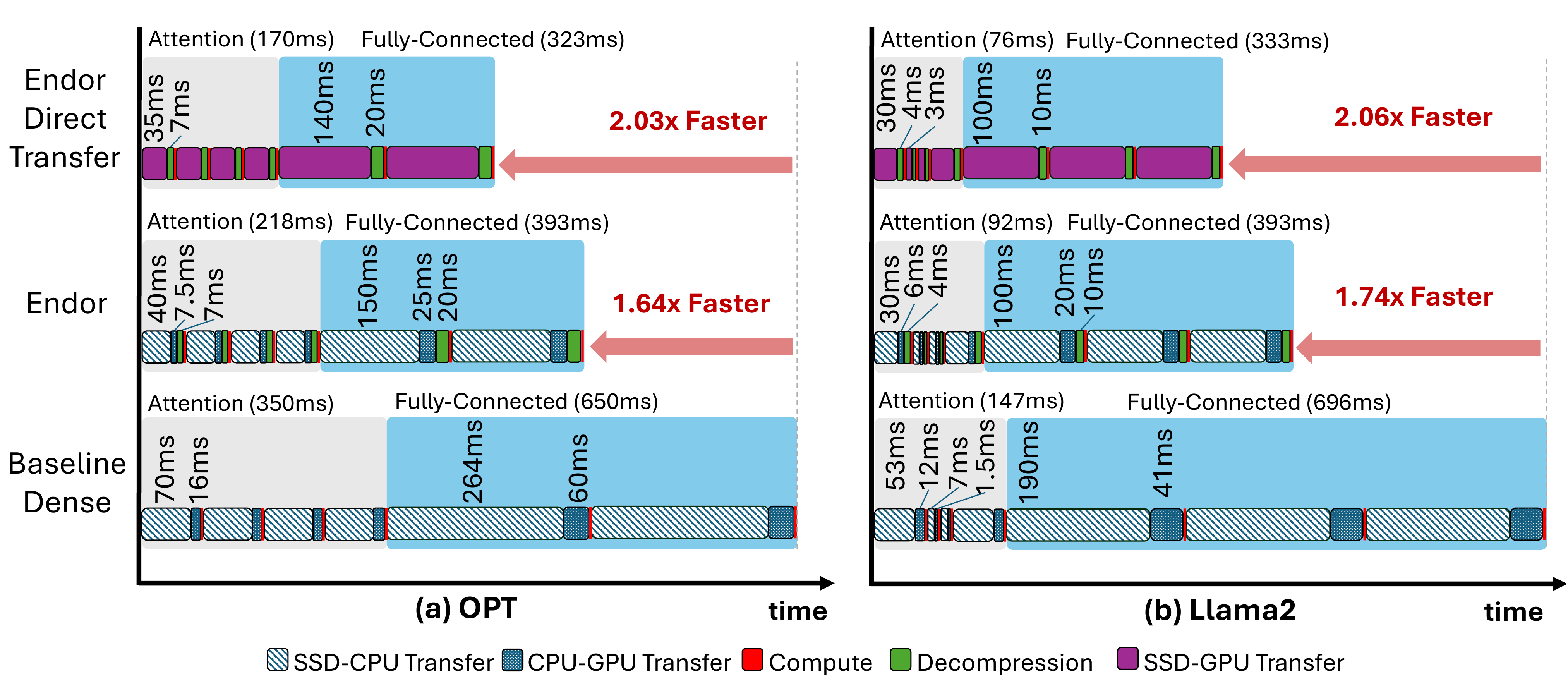}}
\vspace{-10pt}
\caption{Timeline Comparison of SSD-mapped (a) OPT Layer and (b) Llama2 Layer.}
\label{fig:llama-opt-mix}
\vspace{-10pt}
\end{figure}

In Figure~\ref{fig:llama-opt-mix}~a, we present a holistic comparison of each method for one OPT layer, which consists of the linear operations previously discussed. Using direct transfer, an SSD-offloaded OPT layer sees a 2.03$\times$ speedup from a dense OPT layer. Figure~\ref{fig:llama-opt-mix}~b shows how the cumulative speedup is achieved in an offloaded Llama Layer.

\subsection{Direct SSD-GPU Transfer}
\label{sec:gpudirect}

As decompression is done on GPU, SSD-offloaded weights can be read directly into GPU via Direct Memory Access for supported GPUs. We utilize NVIDIA GPUDirect Storage to implement direct SSD-GPU transfer. 

\begin{figure}[b]
\centerline{\includegraphics[width=1.0\textwidth]
{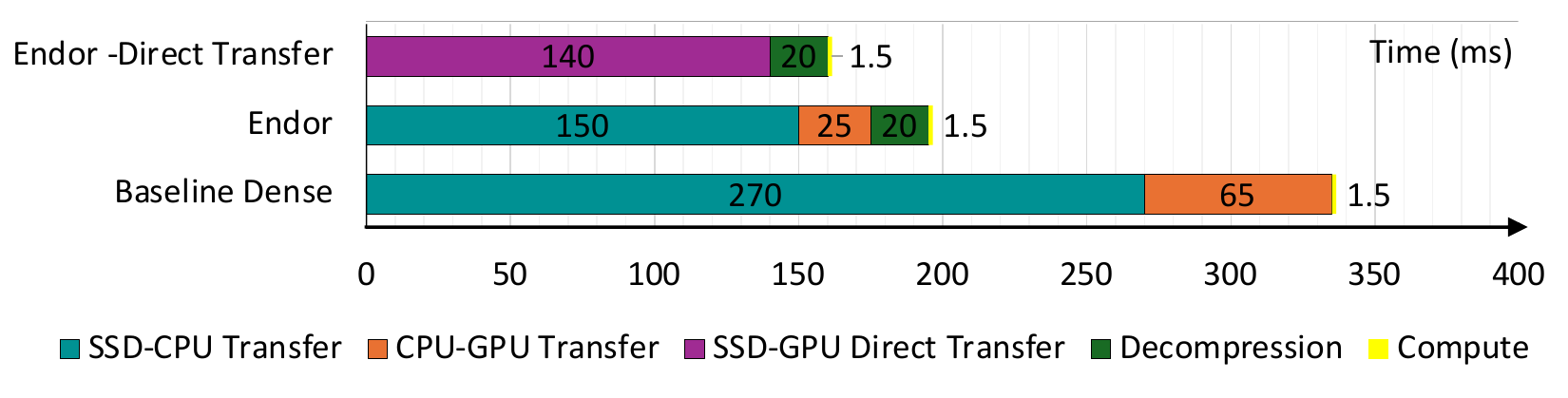}}
\caption{Timeline comparison of a fully-connected operation with SSD-GPU direct transfer}
\label{fig:fig7}
\end{figure}

As shown in Figure~\ref{fig:fig7}, weight transfer latency of an SSD-offloaded linear operation is further reduced. Instead of a two step transfer from SSD to CPU then from CPU to GPU, weight is loaded with a single transfer through the PCIe bus. This achieves a similar to slightly faster bandwidth compared to SSD-CPU transfer, effectively removing the CPU-GPU transfer latency. End-to-end, a single pass through OPT-66B took 24.2s. This is a 1.29$\times$ additional speedup from Endor, and an overall 2.23$\times$ speedup from naive offloaded inference. We include additional measurements of direct SSD-GPU transfer in Appendix~\ref{sec:appendix_GPUDirect}. 

\subsection{Accuracy}

Because Endor sparse format retains the exact non-zero values and the positions of the pruned weight matrix, it preserves the performance of the utilized pruning method. For our workload OPT-66B pruned with SparseGPT~\cite{sparsegpt}, we measured 9.34 perplexity on the WikiText~\cite{wikitext} validation set, which was very close to 9.33 perplexity measure with the dense model. When 2:4 structured sparsity is enforced, we measure the deteriorated perplexity of 10.07, highlighting the value of compressing unstructured sparsity. 
While we employed SparseGPT~\cite{sparsegpt} to prune OPT-66B, we use Wanda to prune Llama2-70B. Dense baseline achieved 3.12 perplexity on WikiText dataset. Unstructured pruning achieved 3.97 perplexity, while 2:4 sparsity achieved 5.20 perplexity.  

\subsection{Joint Application with Quantization}
\label{sec:quantization}

\begin{figure}[htbp]
\centerline{\includegraphics[width=0.90\textwidth]
{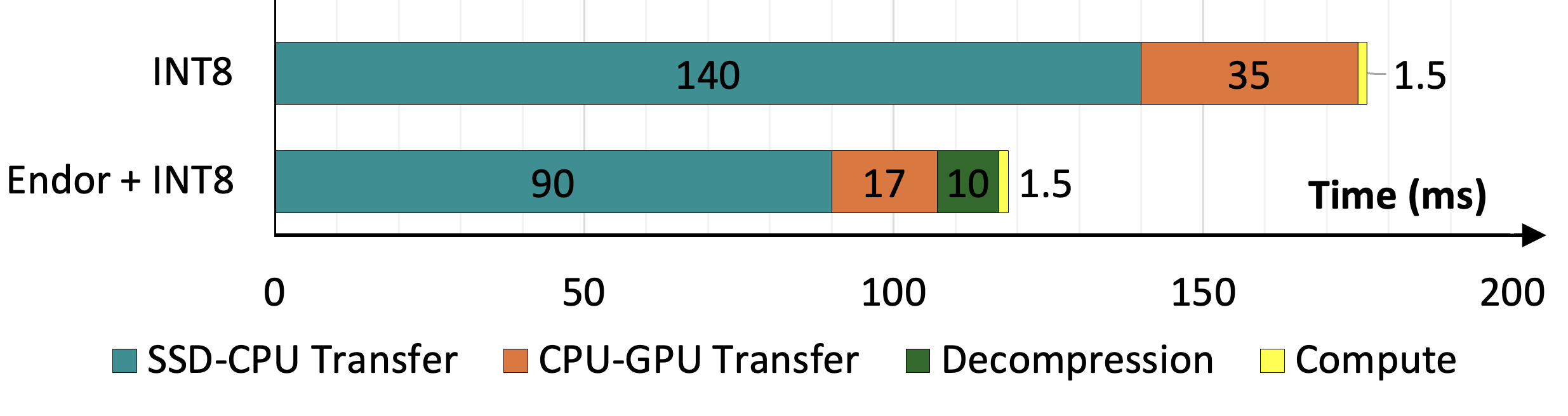}}
\caption{Timeline comparison of a fully-connected operation with INT8 quantization}
\vspace{-10pt}
\label{fig:fig8_final}
\end{figure}

To validate Endor's applicability, we jointly employ Endor on an offloaded inference with SmoothQuant~\cite{smoothquant} 8-bit quantization, using the pruning of Wanda~\cite{wanda}. Figure~\ref{fig:fig8_final} shows that Endor effectively reduces the weight transfer latency to achieve a 1.48$\times$ speedup. Note that the reduction of weight transfer latency is smaller compared to float16, reflecting the reduced compression ratio.

\subsection{Extent of Support}
The current implementation of Endor supports OPT-66B~\cite{opt} and Llama2-70~\cite{llama2}. While it offers pruning implementation of SparseGPT~\cite{sparsegpt} and ~\cite{wanda}, weight externally pruned with any pruning mechanism is also supported. It supports any sparsity pattern and sparsity ratio. It can be easily extended to support various LLMs available in the Huggingface Hub. We plan to open source Endor to stimulate a broader usage of LLM offloaded inference.

\section{Conclusions}
This work analyzed the offloaded inference of LLMs, a crucial solution to the democratization of LLMs that enables execution on constrained platforms. We identified the weight transfer between storage and compute elements as the bottleneck in end-to-end latency. We proposed a sparse format that compressed pruned LLM weights to reduce the memory footprint, effectively reducing the weight transfer overhead with minimal decompression overhead. We showed that our sparse format compression can be applied to existing optimization methods such as quantization and activation sparsity for maximum speedup. This solution can be applied on production-level constrained platforms such as phones, robots, and personal assistants. Additionally, this line of work reduces the weight transfer overhead to the scale of computation, therefore allowing computation and weight transfer to overlap. This paves the way for a complete overlap of computation with weight transfer that will result in the offloaded inference latency match a naive model-in-GPU inference latency.  

\section{Discussion and Future Work}

\textbf{Joint Application of Multiple Optimizations.} We evaluated Endor's compatibility with different optimization methods in terms of speedup. Further studies must be done to ensure that joint application of multiple optimizations retains the accuracy of dense LLMs. As Endor retains the practicality of unstructured sparse patterns for LLM pruning, we hope Endor stimulates robust research in efficient LLM deployment.  

\textbf{Offloading to Multiple SSDs.} Weight transfer latency from SSD is constrained to the read bandwidth of the SSD. While the fastest SSDs provide read bandwidths up to 7000 MB/s, it is far below the achievable read bandwidth of both CPU and PCIe bus even after utilizing SSD-GPU direct transfer. Higher bandwidth is achievable by utilizing multiple SSDs. As Endor decompression is easily parallelizable, an efficient mapping of LLM weights to multiple SSDs will significantly reduce the weight transfer latency.  

\textbf{Efficient Compute Unit.} While our sparse format effectively reduces the amount of data transferred, its computation on GPU is done in a full dense matrix fashion, overlooking the possible reduced amount of computation. Nvidia GPUs support the computation of fine-grained sparsity but lack the support for a fully unstructured sparsity. When valid computations are correctly identified, the same computation throughput can be achieved with a smaller number of compute elements.

\newpage
\printbibliography
\newpage
\appendix

\section{Endor Per-Operation Breakdown}
\label{sec:otherops}

Offloaded operations in OPT-66B can be categorized by the weight matrix shape. In Section~\ref{sec:perlayerspeedup}, we discussed the speedup on fully-connected operation with weight matrix $9,216\times36,864$. In this section, we provide additional measurements of the other offloaded operations. First category include four linear operations in the attention sub-layer: key projection, query projection, value projection, and output projection. Each operation is a weight matrix of size $9,216\times9,216$. Second category is the other linear operation within the fully-connected sub-layer with a weight matrix of size $36,864\times9,216$.  

\begin{figure}[htbp]
\centerline{\includegraphics[width=0.80\textwidth]
{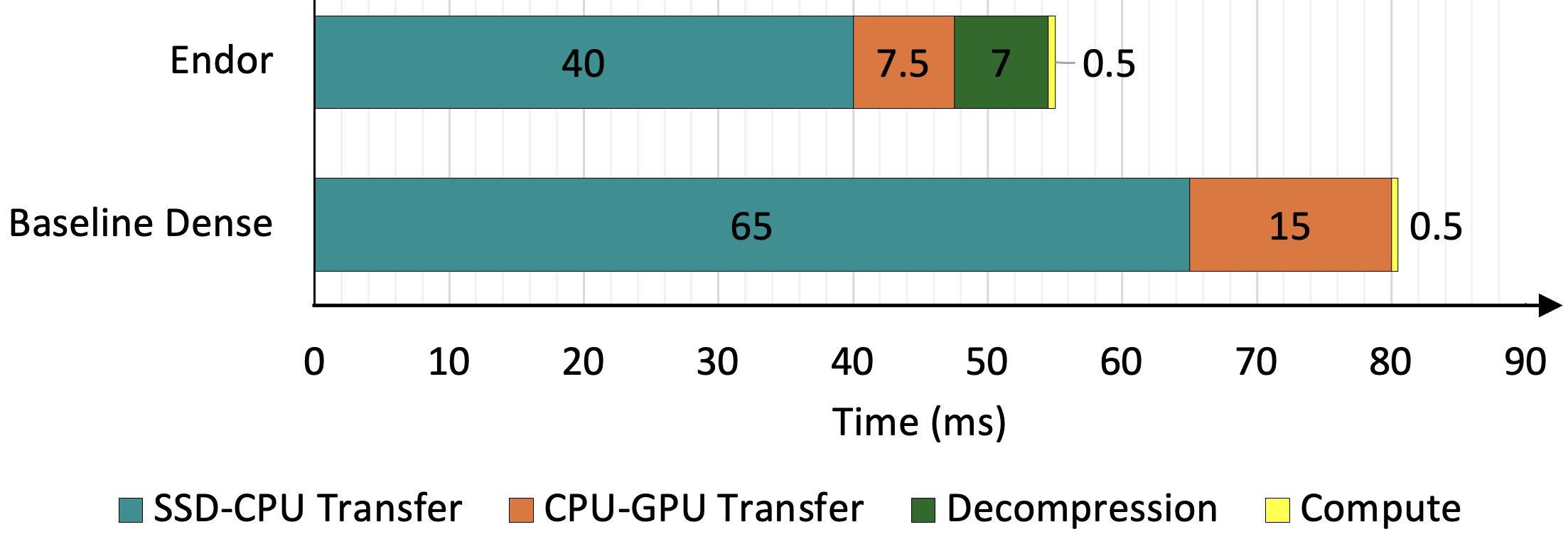}}
\caption{Timeline of SSD-mapped attention linear operation (weight size: $9216\times9216$).}
\label{fig:fig9}
\end{figure}

\begin{figure}[htbp]
\centerline{\includegraphics[width=0.80\textwidth]
{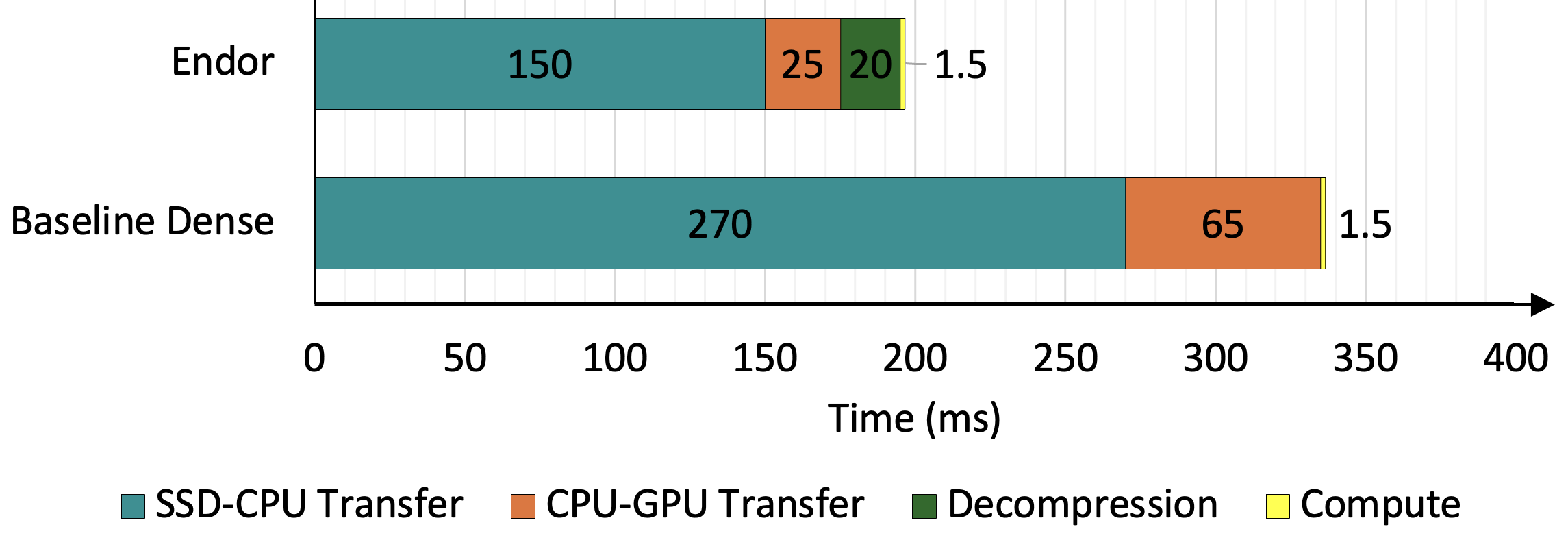}}
\caption{Timeline of SSD-mapped fully-connected linear operation (weight size: $36864\times9216$).}
\label{fig:fig10}
\end{figure}

Figure~\ref{fig:fig9} shows the offloaded inference of the SSD-mapped linear operation from the attention sub-layer. Compressing the weight with Endor sparse format yields the reduction of weight transfer latency near proportionate to the compression ratio with minimal decompression overhead performed on GPU. Figure~\ref{fig:fig10} shows the offloaded inference of the SSD-mapped linear operation from fully-connected sub-layer. Both linear operations inside the fully-connected sub-layer, each with weight matrix sized $9,216\times36,864$ and $36,864\times9,216$ shows the same latency. 

\section{Direct SSD-GPU Transfer}
\label{sec:appendix_GPUDirect}

\begin{figure}[htbp]
\centerline{\includegraphics[width=0.90\textwidth]
{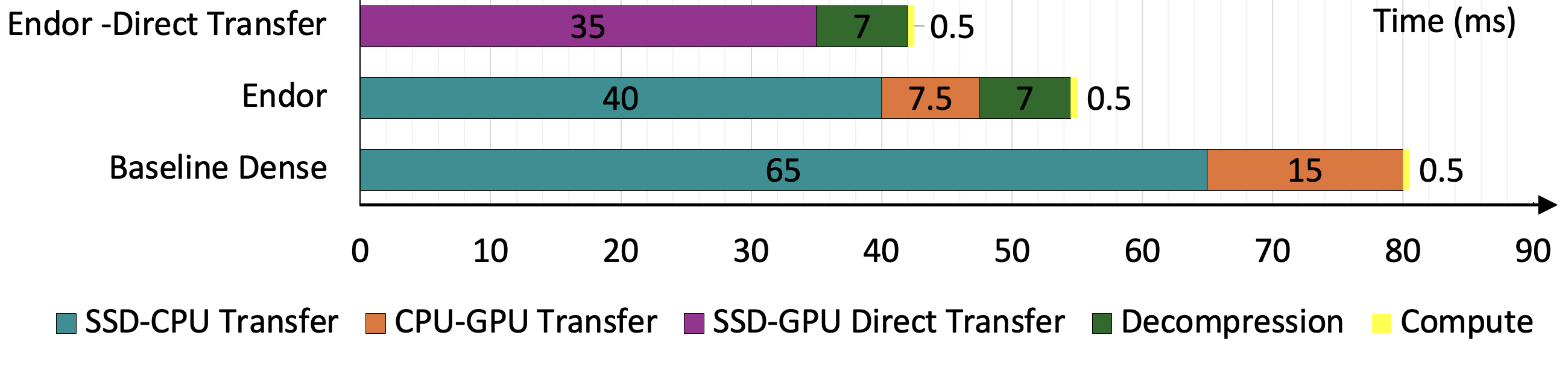}}
\caption{Timeline of SSD-mapped attention linear operation using SSD-GPU Transfer.}
\label{fig:fig11}
\end{figure}

To accompany the measurements of Section~\ref{sec:gpudirect}, we provide additional measurements of applying direct SSD-GPU transfer on offloaded operation from the attention sub-layer. Figure~\ref{fig:fig11} shows that directly loading offloaded weight into GPU bypassing CPU removes the latency of CPU-GPU transfer.

\section{Endor on Llama2-70B}
\label{sec:appendix_Llama}
\subsection{Workload Specification}

We validate Endor sparse format for Llama2-70B in float16 precision. Llama2-70B consists of 80 Llama layers. Each Llama layer is divided into an attention sub-layer and a fully-connected sub-layer. The attention sub-layer contains four linear operations: key projection, query projection, value projection, and output projection. However, one difference from OPT-66B is that Llama uses group query attention. While query projection and output projection uses weight matrix of size $8,192\times8,192$, key projection and value projection uses weight matrix of size $1,024\times8,192$. Fully-connected sub-layer contains three linear operations. The two upward projections (gate projection and up projection) use weight matrix of size $8,192\times28,672$ and one downward projection uses weight matrix of size $28,672\times8,192$. For offloaded inference, we mapped 6 Llama layers on GPU, 10 layers on CPU, and 64 layers on SSD.

\subsection{Per-Operation  Speedup}
\begin{figure}[htbp]
\centerline{\includegraphics[width=0.90\textwidth]
{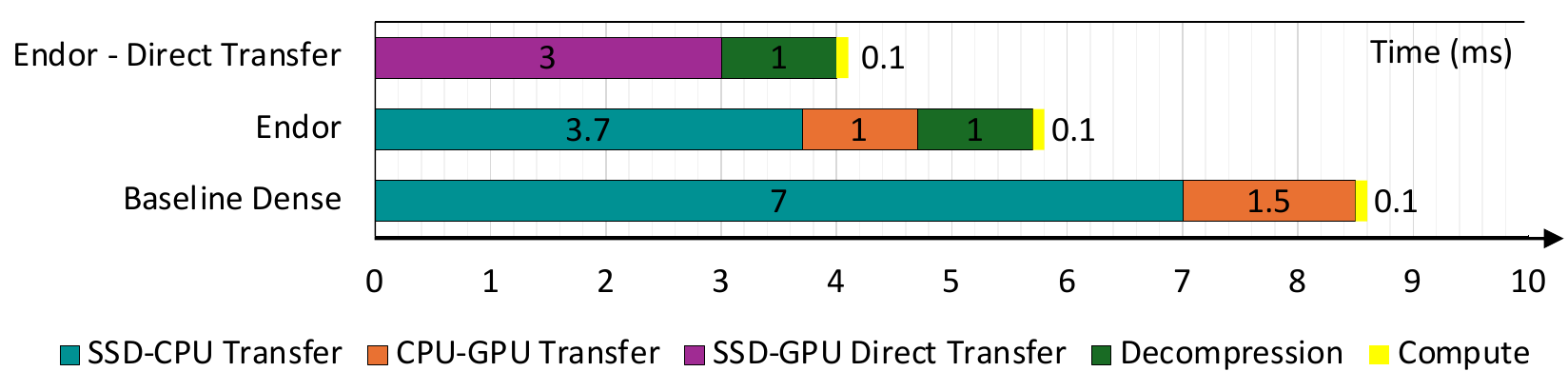}}
\caption{Timeline of SSD-mapped attention linear operation (weight $1,024\times8,192$).}
\label{fig:fig13}
\end{figure}

\begin{figure}[htbp]
\centerline{\includegraphics[width=0.90\textwidth]
{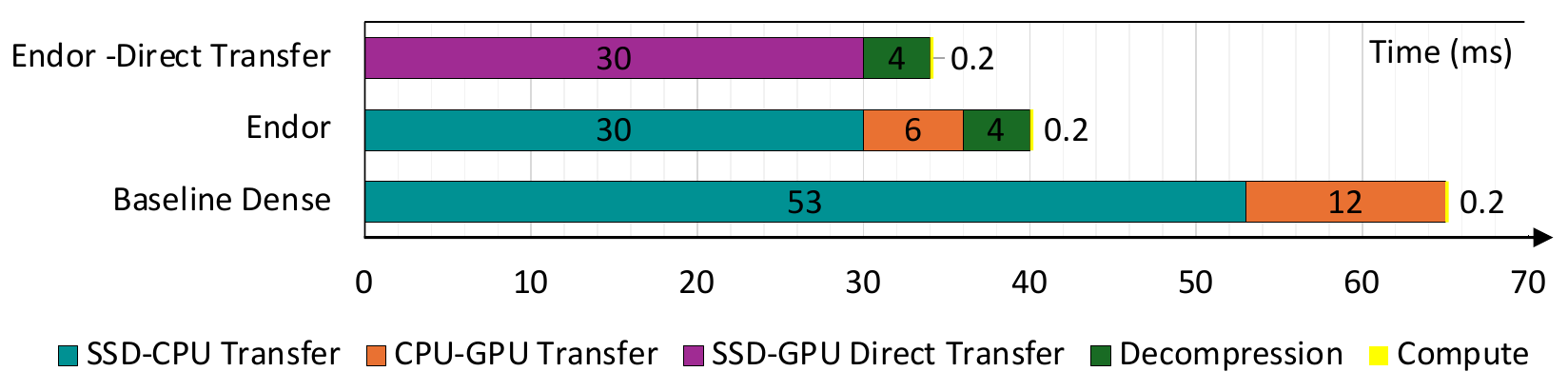}}
\caption{Timeline of SSD-mapped attention linear operation (weight $8,192\times8,192$).}
\label{fig:fig14}
\end{figure}

\begin{figure}[htbp]
\centerline{\includegraphics[width=0.90\textwidth]
{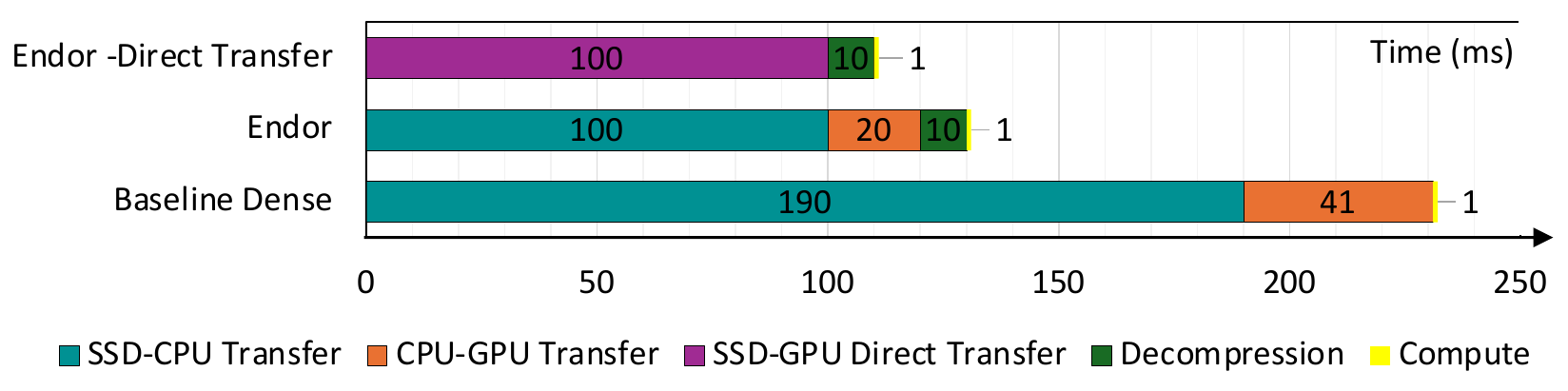}}
\caption{Timeline of SSD-mapped fully-connected linear operation (weight $28,672\times8,192$).}
\label{fig:fig15}
\end{figure}

We present Endor's effect on the three categories of linear operations, of weight matrix size $1,024\times8,192$, $8,192\times8,192$, and $28,672\times8,192$. This time, we present the timeline of dense baseline, Endor, and Endor with direct SSD-GPU transfer. While the weight latency differs with the size of weight transferred, the effect of Endor is evident in all operation in both attention sub-layer (Figure~\ref{fig:fig13}, Figure~\ref{fig:fig14}) and fully-connected sub-layer (Figure~\ref{fig:fig15}). Endor effectively tackles the dominant weight transfer latency of offloaded inference by half with minimal decompression overhead. With GPU support, SSD-GPU direct transfer further reduces the latency by combining the SSD-CPU and CPU-GPU weight transfer into a single transaction. 

\subsection{End-to-End Speedup}
We measured the execution time for single pass through Llama2-70B during text generation. For accurate comparion with dense offloaded inference, we kept the same device mapping: Layers 0 to 5 on GPU, layers 6 to 15 on CPU, and layers 16 to 79 on SSD. 

\begin{figure}[htbp]
\centerline{\includegraphics[width=0.8\textwidth]
{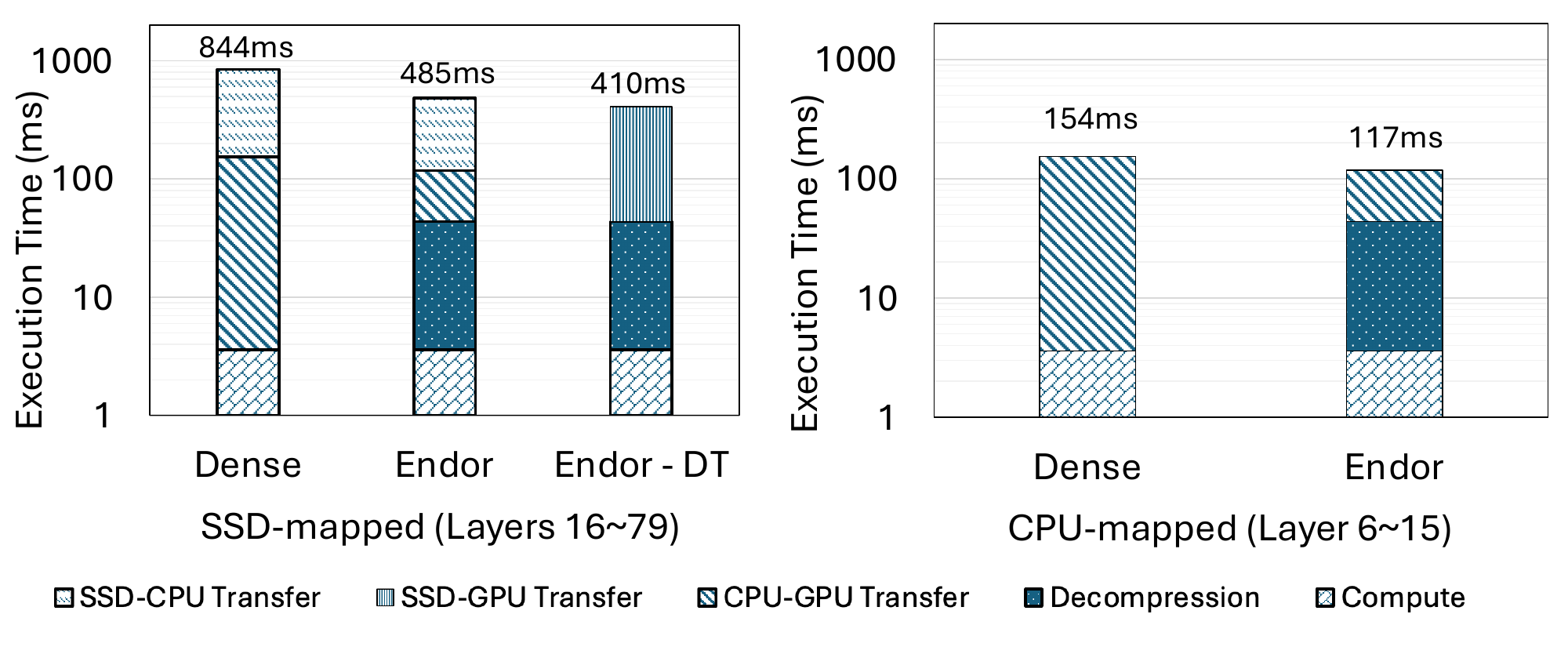}}
\caption{Execution time comparison of dense and Endor offloaded Llama layers for SSD-mapped layers (on left) and CPU-mapped layers (on right).}
\label{fig:17}
\end{figure}

For a single pass through Llama2-70B, dense offloaded inference took 57s while Endor offloaded inference took 35.2s, an overall 1.62$\times$ speedup. Figure~\ref{fig:17} breaks down the speedup into each device-mapped layers. Note that the latency of GPU-mapped layers remain the same. Using direct SSD-GPU transfer, a single pass took 27s, an overall 2.11$\times$ speedup. 

Weight compression with enables more layer weights to reside on CPU. Under the same DRAM footprint, now 20 layers can be CPU-mapped. For a single pass through Llama2-70B, this took 32s without direct transfer and 24s with direct transfer, each an overall 1.78$\times$ and 2.37$\times$ speedup.

\newpage

\end{document}